\documentclass{article}
\usepackage{spconf,amsmath,graphicx,hyperref}
\usepackage{xcolor}
\usepackage{soul}
\usepackage{booktabs}
\usepackage{makecell}
\usepackage{amssymb}
\usepackage{array}
\usepackage{pifont}
\usepackage[normalem]{ulem} 
\usepackage{caption}
\usepackage{balance}   
\usepackage{placeins}

\title{QECGBERT: A Scalable ECG Foundation Model Based on Self-Supervised Language Learning via VQ-VAE Waveform Tokenization}

\name{Aruna Mohan$^{\star}$ \qquad
      Burcu Ozek$^{\star}$\qquad
      Or Zilbershot$^{\star}$ \qquad
      Fatemeh Afghah$^{\star \dagger}$}
\address{$^{\star}$Walkky LLC, Boston, MA, USA \\
         $^{\dagger}$Clemson University, Clemson, SC, USA \\
         \{aruna, burcu, or, fatemeh\}@walkky.com}

\begin{document}

%
\maketitle
\begin{abstract}
AI foundation models play a pivotal role in advancing heart disease diagnosis from electrocardiogram (ECG) data, particularly for rare diseases where the lack of sufficient labeled data hinders supervised learning. However, most self-supervised learning methods treat ECGs as generic sequence or image data without exploiting underlying physiological structure. To address this limitation, we propose QECGBERT, a self-supervised ECG language model that explicitly encodes physiological semantics by tokenizing the P, QRS, T, and isoelectric background waveforms. Vector-quantized variational autoencoder (VQ-VAE) tokenization is applied to each 12-lead ECG, with tokens enriched by continuous waveform features to preserve morphology lost during quantization without increasing sequence length. This design enables compact BERT masked language modeling with short sequences, improving scalability and reducing computational cost relative to long-sequence approaches. QECGBERT applies BERT pre-training to the MIMIC-IV dataset containing over 800,000 unlabeled ECG recordings, and fine-tuning on labeled PTB-XL, CPSC2018, and Chapman--Shaoxing--Ningbo datasets for evaluation. The results demonstrate that QECGBERT achieves comparable or superior performance to published benchmarks, offering a scalable, physiologically aware approach to ECG foundation modeling for cardiac diagnosis.
\end{abstract}

\vspace{-5 pt}
\section{Introduction}
\label{sec:intro}
\vspace{-3 pt}

\begin{figure*}[t]
  \centering
  \includegraphics[width=0.9\textwidth]{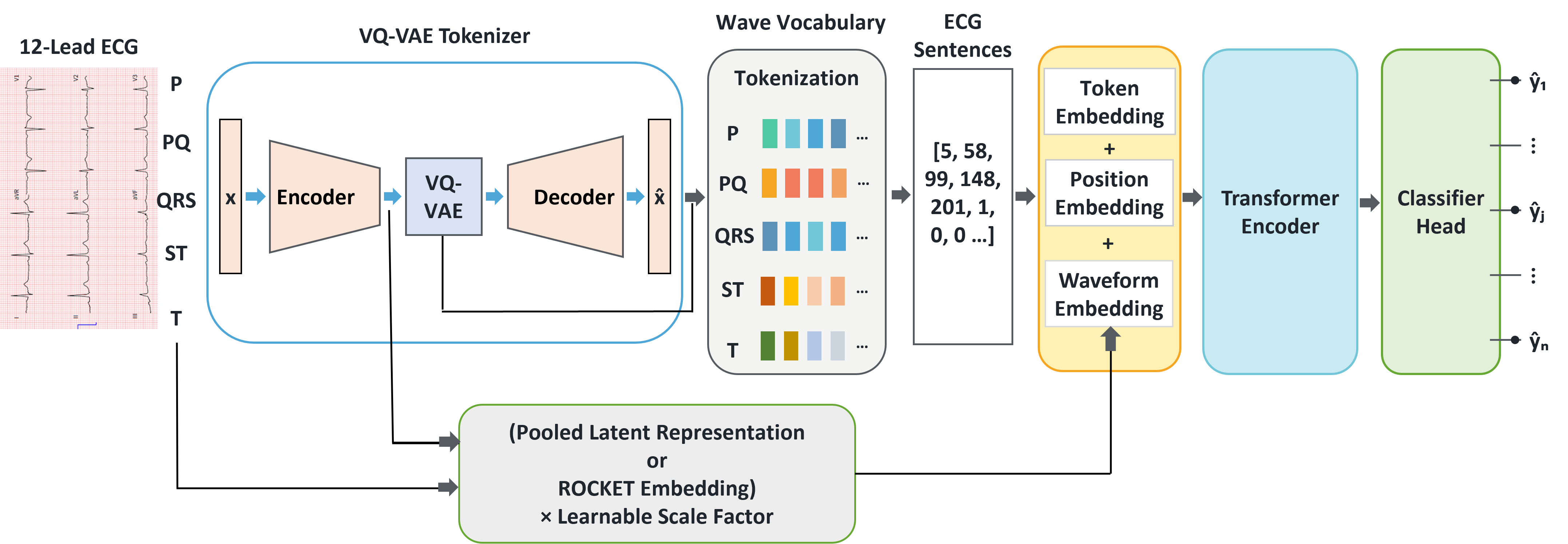}
  \vspace{-5pt}
  \caption{An overview of our proposed QECGBERT model.}
  \label{fig:model}
  \vspace{-10pt}
\end{figure*}

Electrocardiograms (ECG) are a crucial non-invasive tool for monitoring and diagnosing cardiovascular disease. AI foundation models trained on ECG signals are increasingly vital for enabling large‑scale diagnostic support \cite{song2024foundation}. Their impact is particularly significant for rare cardiac conditions, where the paucity of labeled data constrains fully supervised approaches and underscores the need for self‑supervised pretraining on large unlabeled ECG corpora to learn robust physiological representations.

ECG self-supervised learning (SSL) methods generally utilize contrastive or generative approaches. Contrastive methods \cite{liu2023contrastive} learn discriminative ECG features by distinguishing between positive and negative pairs, though poor pair selection may obscure key morphology \cite{lan2023badpair}. The recent ECG-CPC model applied contrastive predictive coding to state-space models, better capturing temporal dependencies \cite{ecgcpc2026iclr}. Multimodal contrastive learning has shown promise in zero-shot ECG classification \cite{liu2024merl, wang2025melp}. However, these models are predicated on textual ECG reports, which are often sparse.

Generative ECG SSL models learn by reconstructing original signals from their masked versions \cite{sawano2024, zhou2023masked}. Recent examples include ST-MEM \cite{na2024stmem}, which patchified ECGs and reconstructed masked patches for 12-lead sequences, and HeartLang \cite{heartlangjin2025}, which tokenized heartbeats based on QRS waveforms to generate 12-lead sequences. These models relied on long, costly sequences for masked token prediction. More recently, ECG2TOK \cite{ecg2tok} has demonstrated strong performance via self-distillation and online clustering of patchified 12-lead ECG data for masked sequence modeling. A different approach pioneered by ECGBERT \cite{choi2023ecgbert} clustered P, QRS, T, and background waveforms from ECGs into vocabulary tokens. RhythmBERT \cite{rhythmbert} applied clustering to latent representations rather than raw waveforms to improve scalability, using lead II only. However, RhythmBERT decoupled clustering from autoencoder training, and its reliance on lead II alone left cross-lead morphology unaddressed.

In this work, we propose QECGBERT, a scalable self-supervised ECG language model that explicitly encodes physiological semantics by tokenizing the P, QRS, T, and isoelectric background waveforms of 12-lead ECGs, pre-trained on the MIMIC-IV dataset containing over 800,000 unlabeled ECGs. Our contributions are: (1) We apply vector-quantized variational autoencoder (VQ-VAE) tokenization \cite{oord2017neurips} to each waveform, jointly learning the encoder and discrete codebook for each 12-lead segment type. This yields a compact, physiologically grounded wave vocabulary. (2) We enhance the discrete token embeddings with additional embeddings that capture intra-cluster variations, using either the latent waveform representations from the trained segment encoder or the untrained ROCKET model \cite{rocket}. These complementary features retain waveform differences lost during quantization without increasing sequence length. (3) We pre-train a BERT transformer via masked language modeling (MLM) over short waveform sequences, improving scalability and computational efficiency compared to long-sequence modeling. (4) We evaluate QECGBERT on multi-labeled PTB-XL, CPSC2018, and Chapman--Shaoxing--Ningbo datasets, demonstrating performance comparable or superior to published benchmarks despite using short sequences.

\vspace{-5pt}
\section{Methodology}
\label{sec:methodology}
\vspace{-3pt}

The architecture of QECGBERT is depicted in Fig.~\ref{fig:model}. The model utilizes P, QRS, T, and background waves delineated from 12-lead ECG data, preserving spatial information across leads. The background isoelectric segments highlight ST and PR changes, key to diagnosing myocardial infarction, ischemia, and pericarditis. The TP segment serves only as a baseline reference and is excluded from the model.

\vspace{-5pt}
\subsection{ECG Waveform Tokenizer}
\label{ssec:VQVAE}
\vspace{-3pt}

We applied a 0.5 Hz high-pass 5th order Butterworth filter followed by a 50 Hz notch filter to pre-process the raw ECG signals. Next, ECG delineation was performed by using the discrete wavelet transform (DWT) method \cite{martinez2004wavelet} to segment the ECG into P, QRS, and T waveforms by identifying onsets and offsets based on lead II and using the same starting and ending waveform indices for all leads. The isoelectric intervals between P and QRS waveforms, denoted by PQ, and between QRS and T waveforms, denoted by ST, are treated as interstitial segment types. These five 12-lead segment types serve as inputs to the VQ-VAE tokenizers. The extracted waveforms are z-normalized and truncated or zero-padded to the 99th percentile segment length based on the pre-training dataset for each segment type to achieve a uniform input length for the tokenizer. The corresponding lengths are 88, 88, 176, 128, and 120, respectively, for P, PQ, QRS, ST, and T segments.

The waveform tokenizer consists of one VQ-VAE module per segment type, each tasked with learning a discrete codebook for a specific 12-lead segment type. The 12-lead waveforms constitute the module input, enabling the model to capture lead variations without separate per-lead representations.  Each VQ-VAE module learns a convolutional encoder, a quantized codebook, and a decoder end-to-end. In this process, the encoder maps a waveform to a continuous latent, the VQ-VAE quantizes the latent embedding to its nearest entry in a learned codebook, and the decoder reconstructs the original waveform from the quantized latent:
\begin{eqnarray}
& z_n^\tau = f_\theta^\tau(x_n^\tau), \nonumber \\
& q_n^\tau = e_{k^*}, \ \ k^* = \arg\min_k \lVert z_n^\tau - e_k \rVert_2, \nonumber \\
& \hat{x}_n^\tau = g_\phi^\tau(q_n^\tau),
\end{eqnarray}
where $x_n^\tau \in \mathbb{R}^{12\times L_{\tau}}$ is the $n$-th 12-lead segment of type $\tau$ with per-lead segment length $L_{\tau}$, $f_\theta^\tau$ is the encoder parametrized by $\theta$, $z_n^\tau \in \mathbb{R}^{d_\tau}$ its continuous latent with embedding dimension ${d_\tau}$, $\{e_k\}_{k=1}^{K_\tau}$ the codebook of size $K_\tau$ for wave type $\tau \in \{ \mathrm{P}, \allowbreak \mathrm{PQ}, \allowbreak \mathrm{QRS}, \allowbreak \mathrm{ST}, \allowbreak \mathrm{T} \}$, $q_n^\tau$ the quantized latent, and $g_\phi^\tau$ the decoder parametrized by $\phi$ producing the reconstruction $\hat{x}_n^\tau$. Training minimizes the standard VQ-VAE objective, combining a Huber reconstruction loss $\ell_\beta$ with $\beta=1.0$ and the codebook commitment loss:
\begin{equation}
\mathcal{L}_\mathrm{VQ}^{\tau} =
\underbrace{\ell_\beta\!\left(x_n^\tau - \hat{x}_n^\tau\right)}_{\text{reconstruction}} +
\gamma \underbrace{\lVert z_n^\tau - \mathrm{sg}[q_n^\tau] \rVert_2^2}_{\text{commitment}},
\end{equation}
where $\mathrm{sg}[\cdot]$ denotes the stop-gradient operator and the commitment cost $\gamma=0.25$ weights the commitment term. Codebook entries are updated via an exponential moving average (EMA) of assigned latents with an EMA decay coefficient of $0.99$, with dead-code reinitialization applied periodically to mitigate codebook collapse. The codebook size is 48 with embedding dimension ${d_\tau}=128$ for all segment types. 

Each encoder applies two strided 1D convolutions with a kernel size of 3 and GELU activation to progressively downsample the input by a factor of 4, followed by a shape-preserving convolution and a residual stack that refines the bottleneck representation. The convolutional layers use 64 and 128 channels, while the residual stack has 4 layers with 32 hidden channels. The resulting feature map is flattened, linearly projected into the latent embedding space, and L2-normalized. The decoder mirrors this structure by using a linear layer, residual stack, shape-preserving convolution, and two transposed convolutions to upsample and reconstruct the original segment. The VQ-VAE modules have a total of 6.4M parameters over all 5 segments.

The vocabulary for beat sentence construction is constituted by the per-segment codebook indices offset to form a disjoint index range. These indices serve as discrete tokens, concatenated in physiological order (P--PQ--QRS--ST--T) to form the heartbeat ``sentences'' for the BERT model.

\vspace{-5pt}
\subsection{BERT Model}
\label{ssec:BERT}
\vspace{-3pt}

Our self-supervised learning framework is a transformer encoder pre-trained to capture contextual representations from symbolic beat sentences. Each beat sentence is structured with \texttt{[PAD]} tokens to pad sequences to a uniform length, a \texttt{[SEP]} token marking the end of a sequence, and \texttt{[MASK]} tokens for MLM pre-training. The sentence length is selected randomly to be 1, 2, 6, or 8 consecutive heartbeats from an ECG record to enhance the generalizability of pre-training \cite{choi2023ecgbert}. The resulting maximum sequence length is 41.   

\noindent\textbf{Input Embeddings.} (1) Token embeddings: A trainable embedding matrix maps the discrete token IDs from the input beat sentences to ``word'' vectors of dimension $d_{\text{model}}$. (2) Positional embeddings: Learned positional embeddings are added to each token embedding to incorporate sequence order. (3) Waveform embeddings: The token IDs capture the discrete codebook representations, but not intra-cluster variability among waveforms assigned to a single codebook entry. Following the ECGBERT approach \cite{choi2023ecgbert}, we add extra embeddings extracted directly from the waveforms to capture the intra-cluster variability. We evaluate three model variants:

(3.1) \underline{QECGBERT (plain)} is the baseline without extra waveform embeddings. 

(3.2) \underline{QECGBERT (latent)} adds the pooled feature representation from the output of the residual stack of the already-trained VQ-VAE module encoder, concatenating average and maximum pooling to a feature dimension of 2560. 

(3.3) \underline{QECGBERT (ROCKET)} adds extra features extracted from the waveforms by the untrained ROCKET model \cite{rocket}, a fixed random convolutional feature extractor, employing 1000 kernels randomly sampled from kernel lengths of 7, 9, or 11, with ``same'' or ``valid'' padding, applied to a random subset of leads. Each kernel outputs the proportion of positive values and the maximum after convolution \cite{rocket}, yielding a 2000-dimensional feature vector. 

The extra waveform embeddings are projected to $d_{\text{model}}$ via a trainable linear layer, normalized, scaled by a trainable factor, and added to the token and positional embeddings.

\noindent\textbf{Transformer Encoder.} The encoder is a stack of 8 transformer layers. Each layer consists of a multi-head self-attention sub-layer and a position-wise feed-forward network (FFN) with an inner dimensionality of $4 \times d_{\text{model}}$, with ReLU activation, dropout regularization, and layer normalization.  We use 12 attention heads and $d_{\text{model}}=384$. For MLM pre-training, the encoder layers are followed by linear projection of masked embeddings, GELU activation, layer normalization, and a linear layer mirroring the encoder weights to produce output logits, optimized via cross-entropy loss. Fine-tuning uses mean pooling of embeddings over the sequence length, excluding padding, with a classifier head optimized via multi-label binary cross-entropy loss.

\begin{table*}[t]
\caption{AUROC results for classification on downstream CPSC2018, CSN and CS datasets. For QECGBERT, the error bars represent standard deviations calculated over 9 fine-tuning runs. The best results are highlighted in \textbf{bold}, while the second-best are shown in \colorbox{gray!25}{gray}. Values overlapping within error bars receive the same rank.}
\vspace{-1pt}
\label{tab:results-auroc-c}
\centering
\small
\setlength{\tabcolsep}{4pt}
\resizebox{\textwidth}{!}{
\begin{tabular}{lccc ccc ccc}
\toprule
\textbf{Model} &
\multicolumn{3}{c}{\textbf{CPSC2018}} &
\multicolumn{3}{c}{\textbf{CSN}} & 
\multicolumn{3}{c} {\textbf{CS}} \\
\cmidrule(lr){2-4}\cmidrule(lr){5-7}\cmidrule(lr){8-10}
& \textbf{1\%} & \textbf{10\%} & \textbf{100\%}
& \textbf{1\%} & \textbf{10\%} & \textbf{100\%}
& \textbf{1\%} & \textbf{10\%} & \textbf{100\%} \\
\midrule
ST-MEM (\cite{na2024stmem}) & 
56.69 & 63.32 & 70.39 & \colorbox{gray!25}{59.77} & 66.87 & 71.36 & \textemdash & \textemdash & \textemdash \\
HeartLang (\cite{heartlangjin2025}) &
60.44 & 66.26 & 77.87 & 57.94 & 68.93 & 82.49 & \textemdash & \textemdash & \textemdash \\
RhythmBERT (\cite{rhythmbert}) &
63.72 & 70.09 & 82.38 & \textemdash & \textemdash & \textemdash & 62.48 & 85.36 & 94.11 \\
ECG2TOK (\cite{ecg2tok}) &
\colorbox{gray!25}{68.42} & \colorbox{gray!25}{83.41} & 91.04 & \textbf{72.66} & \textbf{85.14} & \textbf{92.84} & \textemdash & \textemdash & \textemdash \\
\midrule
QECGBERT (plain) & 
\colorbox{gray!25}{66.62 \textpm 5.05} & 80.18 \textpm 0.99 & 87.35 \textpm 0.17 & 53.75 \textpm 2.55 & 76.47 \textpm 0.75 & 83.34 \textpm 0.44 & \colorbox{gray!25}{78.30 \textpm 2.89} &  \colorbox{gray!25}{87.25 \textpm 1.52} & 91.16 \textpm 0.19 \\
QECGBERT (latent) &
\textbf{74.82 \textpm 5.32} & \textbf{86.92 \textpm 0.43} & \colorbox{gray!25}{92.31 \textpm 0.20} & \colorbox{gray!25}{59.08 \textpm 6.38} & \colorbox{gray!25}{80.99 \textpm 1.23} & 89.35 \textpm 0.20 & \colorbox{gray!25}{78.48 \textpm 5.29} & \textbf{90.64 \textpm 0.75} & \colorbox{gray!25}{94.73 \textpm 0.20} \\
QECGBERT (ROCKET) &
\textbf{75.40 \textpm 5.16} & \textbf{87.01 \textpm 0.54} & \textbf{93.02 \textpm 0.17} & 55.97 \textpm 3.03 & \colorbox{gray!25}{82.20 \textpm 1.03} & \colorbox{gray!25}{90.71 \textpm 0.33} & \textbf{83.48 \textpm 1.76} & \textbf{91.52 \textpm 0.48} & \textbf{95.34 \textpm 0.26} \\
\bottomrule
\end{tabular}
}
\vspace{-6pt}
\end{table*}

\begin{table*}[t]
\caption{AUROC results for classification on downstream PTB-XL datasets. As before, the error bars for QECGBERT represent standard deviations calculated over 9 fine-tuning runs. The best results are highlighted in \textbf{bold}, while the second-best are shown in \colorbox{gray!25}{gray}. Values overlapping within error bars receive the same rank.}
\vspace{-1pt}
\label{tab:results-auroc-p}
\centering
\small
\setlength{\tabcolsep}{4pt}
\resizebox{\textwidth}{!}{
\begin{tabular}{l ccc ccc ccc ccc}
\toprule
\textbf{Model} &
\multicolumn{3}{c}{\textbf{PTB-XL Superclass}} &
\multicolumn{3}{c}{\textbf{PTB-XL Subclass}} & 
\multicolumn{3}{c}{\textbf{PTB-XL Form}} & 
\multicolumn{3}{c} {\textbf{PTB-XL Rhythm}} \\
\cmidrule(lr){2-4}\cmidrule(lr){5-7}\cmidrule(lr){8-10}\cmidrule(lr){11-13}
& \textbf{1\%} & \textbf{10\%} & \textbf{100\%}
& \textbf{1\%} & \textbf{10\%} & \textbf{100\%}
& \textbf{1\%} & \textbf{10\%} & \textbf{100\%}
& \textbf{1\%} & \textbf{10\%} & \textbf{100\%} \\
\midrule
ST-MEM (\cite{na2024stmem}) & 
61.12 & 66.87 & 71.36 & 54.12 & 57.86 & 63.59 & 55.71 & 59.99 & 66.07 & 51.12 & 65.44 & 74.85 \\
HeartLang (\cite{heartlangjin2025}) &
\colorbox{gray!25}{78.94} & \textbf{85.59} & 87.52 & \colorbox{gray!25}{64.68} & 79.34 & \textbf{88.91} & \textbf{58.70} & \colorbox{gray!25}{63.99} & \colorbox{gray!25}{80.23} & 62.08 & 76.22 & \colorbox{gray!25}{90.34} \\
RhythmBERT  (\cite{rhythmbert}) &
65.38 & 70.71 & 75.15 & \textemdash & \textemdash & \textemdash & \textemdash & \textemdash & \textemdash & \textemdash & \textemdash & \textemdash \\
ECG2TOK (\cite{ecg2tok}) &
\textbf{81.23} & \textbf{85.68} & \colorbox{gray!25}{87.99} & \textbf{72.31} & \colorbox{gray!25}{79.42} & 84.52 & \colorbox{gray!25}{55.88} & \textbf{73.78} & \textbf{85.28} & \textbf{81.16} & \textbf{89.41} & \textbf{91.41} \\
\midrule
QECGBERT (plain) & 
74.98 \textpm 0.63 & 79.43 \textpm 0.14 & 81.37 \textpm 0.16 & \colorbox{gray!25}{63.23 \textpm 4.45} & 73.98 \textpm 0.96 & 78.84 \textpm 0.59 & 49.92 \textpm 2.04 & 58.08 \textpm 4.36 & 70.37 \textpm 1.56 & 54.28 \textpm 10.32 & 79.53 \textpm 2.36 & 86.84 \textpm 0.44 \\
QECGBERT (latent) &
 \colorbox{gray!25}{79.36 \textpm 1.28} &  \colorbox{gray!25}{83.93 \textpm 0.25} & 86.65 \textpm 0.12 & \textbf{68.50 \textpm 4.64} & \textbf{80.33 \textpm 0.78} &  \colorbox{gray!25}{87.00 \textpm 0.30} & 50.98 \textpm 1.81 &  \colorbox{gray!25}{61.50 \textpm 2.50} & 77.76 \textpm 0.88 &  \colorbox{gray!25}{69.09 \textpm 9.58} & \colorbox{gray!25}{82.20 \textpm 1.05} & 89.29 \textpm 0.40 \\
QECGBERT (ROCKET) &
\textbf{80.73 \textpm 1.42} & \textbf{85.75 \textpm 0.22} & \textbf{88.92 \textpm 0.19} & \textbf{70.87 \textpm 1.94} & \textbf{81.55 \textpm 0.68} & \textbf{88.43 \textpm 0.71} & 51.97 \textpm 2.12 & \colorbox{gray!25}{64.77 \textpm 2.15} & 77.06 \textpm 1.55 & \colorbox{gray!25}{70.83 \textpm 3.89} & \colorbox{gray!25}{83.49 \textpm 0.98} & 89.03 \textpm 0.28 \\
\bottomrule
\end{tabular}
}
\vspace{-10pt}
\end{table*}

\vspace{-5pt}
\section{Experiments and Analysis}
\label{sec:experiments_analysis}

\vspace{-5pt}
\subsection{Pre-training Configuration}
\vspace{-3pt}

\noindent\textbf{MIMIC-IV-ECG.} This publicly available dataset \cite{gow2023mimic}, comprising 800,035 12-lead ECGs sampled at 500 Hz for 10 s durations from 161,352 patients, is used for pre-training. 

\noindent\textbf{Implementation.} Pre-training consists of two sequential stages: VQ-VAE tokenization and BERT MLM pre-training. The first stage trains per-segment VQ-VAE modules, as described in Section \ref{ssec:VQVAE}. The second stage pre-trains the transformer encoder via a masked prediction objective \cite{devlin2019bert} by dynamically masking 15\% of input tokens, employing the paradigm of 80\% masking/10\% random replacement/10\% unchanged tokens \cite{liu2019roberta}. Optimization uses AdamW with a weight decay of $1\times 10^{-4}$ and $3\times 10^{-4}$, peak learning rate of $5\times 10^{-4}$ and $1\times 10^{-4}$, and 60--140 epochs, in the first and second stages, respectively, with cosine annealing, warmup, gradient clipping, and batch size of 256. Both stages employ 85\%/15\% patient-level splits for training and validation.

\vspace{-5pt}
\subsection{Downstream Tasks Configuration}
\vspace{-3pt}
We assess the performance of QECGBERT on datasets used in prior benchmark studies, including single-labeled and multi-labeled records spanning over 100 cardiac conditions. 

\noindent\textbf{PTB-XL.} This public dataset contains 21,837 12-lead ECGs collected from 18,885 patients, sampled at 500 Hz for 10 s \cite{wagner2020ptb}. The classification task has four subsets: Superclass (5 classes), Subclass (23 classes), Form (19 classes), and Rhythm (12 classes). We  follow the official data split for training, validation, and testing \cite{strodthoff2020deep}.

\noindent\textbf{CPSC2018.} This public dataset contains 6,877 12-lead ECGs sampled at 500 Hz for durations from 6 to 60 seconds, with 9 classes \cite{liu2018cpsc}. We follow the MERL split \cite{liu2024merl} to split the data into 70\%, 10\%, and 20\% for training, validation, and testing.

\noindent\textbf{Chapman--Shaoxing--Ningbo (CSN).} This public dataset contains 45,152 12-lead ECG records sampled at 500 Hz for 10 s \cite{zhengCSN}. Following the MERL configuration \cite{liu2024merl}, we remove ECGs with ``unknown'' labels, resulting in a refined dataset of 23,026 ECG records with 38 classes. In place of CSN, RhythmBERT \cite{rhythmbert} used the simpler Chapman--Shaoxing (CS) subset containing 10,646 single-labeled ECGs with 11 classes \cite{zhengcs}. We use both CSN and CS, and split the datasets into 70\%, 10\%, and 20\% for training, validation, and testing, using the MERL split for CSN \cite{liu2024merl}, and a patient-level class-stratified random split for CS.

\noindent\textbf{Implementation.} For downstream classification, we fine-tune the pre-trained BERT model using Low-Rank Adaptation (LoRA, $r=8, \ \alpha=16$) \cite{hu2022lora}, with a fully-trained classifier head. This process tunes about 600,000 parameters, or roughly $4\%$ of the total 15M model parameters. The beat sentences are generated in the same manner as pre-training, and majority-voting is applied to predictions for sequences from the same ECG record. To assess performance under low-resource conditions, we conduct fine-tuning using 1\%, 10\%, and 100\% of the training data, while the validation and test sets are unchanged. We use 3 splits of the training data to generate the 1\% and 10\% training sizes, and run fine-tuning with 3 seeds for each split, yielding 9 runs for each result. For the 100\% training size, the 9 runs only reflect seed variability and non-determinism in parallel computing. Optimization follows the same approach as pre-training. For all downstream tasks, we report the mean and standard deviation of macro-AUROC from the 9 fine-tuning runs. 

\vspace{-5pt}
\section{Results}
\label{sec:results}
\vspace{-3pt}

The macro-AUROC results for our QECGBERT models are shown in Tables \ref{tab:results-auroc-c} and \ref{tab:results-auroc-p}. We compare our results to recent SSL benchmarks trained on ECG data, but exclude MERL \cite{liu2024merl} and MELP \cite{wang2025melp}, which are multimodal models dependent on textual data, and ECG-CPC trained on over 10M ECG records \cite{ecgcpc2026iclr}, which does not show low-resource performance for our downstream datasets. The relevant benchmarks for comparison are ST-MEM \cite{na2024stmem}, HeartLang \cite{heartlangjin2025}, RhythmBERT \cite{rhythmbert}, and ECG2TOK \cite{ecg2tok}. The results demonstrate that QECGBERT is the best-performing for CPSC2018, PTB-XL Superclass, PTB-XL Subclass, and CS, and second-best for CSN. For PTB-XL Form and Rhythm datasets, QECGBERT performs second-best at 10\% training size, and lags by only 1-2\% behind ECG2TOK and HeartLang for Rhythm at 100\% training size. The Form labels, however, describe qualitative morphological and amplitude changes, which are not captured well by our current approach of tokenizing z-normalized waveforms extracted from raw ECG data. This limitation will be addressed in future studies.

Our results show large error bars at 1\% training size, as expected due to training on about 100--200 records under class imbalance conditions. Low-resource SSL is typically the most relevant at 1000s of training records, corresponding to 10\% training size for these datasets. Under these realistic conditions of limited labeled data, QECGBERT consistently shows the best or second-best performance, establishing its utility for low-resource ECG diagnosis. 

Our approach relies on short sequences composed of 1, 2, 6, or 8 randomly selected heartbeats from an ECG, where each heartbeat is represented by 5 waveform tokens, yielding unpadded sequence lengths of 6, 11, 31 or 41. This multi-scale design enables the model to learn both local morphology and global rhythmic patterns via MLM pre-training on compact sequences. In contrast, HeartLang employed longer 12-lead sequences of length 256 \cite{heartlangjin2025}. Although ECG2TOK achieved efficient convergence using 50 patches per 10 s, 500 Hz ECG and aggressive masking \cite{ecg2tok}, it used 12-layer transformer backbones with hidden dimension of 768 for context and target encoders, with an estimated 172M parameters (86M per encoder, assuming ViT-Base scale). Our model has 21.4M parameters (6.4M for VQ-VAE and 15M for BERT), and demonstrates comparable performance using shorter heartbeat-based sequences. This compact design highlights the feasibility of lightweight training and inference, underscoring the efficiency and scalability of our method.

\vspace{-5pt}
\section{Conclusions}
\label{sec:conclusions}
\vspace{-3pt}

This work introduces QECGBERT, an ECG SSL framework leveraging VQ-VAE waveform tokenization of 12-lead ECGs, with added waveform features to retain morphology differences lost during quantization. This compact, physiologically grounded tokenization approach enables scalable and efficient short-sequence MLM pre-training. QECGBERT achieves AUROC performance competitive with or surpassing prior long-sequence benchmarks, offering a scalable SSL pathway for ECG-based diagnostics.

\FloatBarrier      
\clearpage         
\balance
\bibliographystyle{IEEEbib}
\ninept
\bibliography{strings,refs}

\noindent \textbf{Conflict of Interest Disclosure:} No funding was received for conducting this study. Nothing to disclose.\\
\textbf{Compliance with Ethical Standards:} This is a numerical simulation study for which no ethical approval was required.

\end{document}